\documentclass{article}
\usepackage{enumitem}

\usepackage[square,sort,comma,numbers]{natbib}
\usepackage{amsmath}
\usepackage{graphicx}
\usepackage{arxiv}
\usepackage{setspace}
\usepackage[utf8]{inputenc}
\usepackage[T1]{fontenc}
\usepackage{hyperref}
\usepackage{url}
\usepackage{booktabs}
\usepackage{amsfonts}
\usepackage{nicefrac}
\usepackage{microtype}
\usepackage{lipsum}
\usepackage{mathtools}
\usepackage{epigraph}
\usepackage{csquotes}

\usepackage{tikz}
\usetikzlibrary{arrows.meta,positioning,fit,backgrounds,shadows.blur}

\usepackage{pgfplots}
\usepgfplotslibrary{colormaps}
\usepgfplotslibrary{colorbrewer}
\usepackage{sansmath}
\pgfplotsset{compat=1.18}

\date{}

\title{Testing the Machine Consciousness Hypothesis}

\author{
\bf Stephen Fitz\thanks{This work forms part of an ongoing research collaboration at CIMC (\url{https://cimc.ai}). For correspondence and inquiries: \texttt{mail@stephenfitz.net}}
\\ \textsc{California Institute for Machine Consciousness}
}

\begin{document}

\maketitle
\onehalfspacing

\begin{abstract}
The \emph{Machine Consciousness Hypothesis} states that consciousness is a substrate-free functional property of computational systems capable of \emph{second-order perception}. I propose a research program to investigate this idea in silico by studying how \emph{collective self-models} (coherent, self-referential representations) emerge from distributed learning systems embedded within universal self-organizing environments. The theory outlined here starts from the supposition that consciousness is an emergent property of collective intelligence systems undergoing synchronization of prediction through communication. It is not an epiphenomenon of individual modeling but a property of the language that a system evolves to internally describe itself. For a model of base reality, I begin with a minimal but general computational world: a cellular automaton, which exhibits both computational irreducibility and local reducibility. On top of this computational substrate, I introduce a network of local, predictive, representational (neural) models capable of communication and adaptation. I use this \emph{layered model} to study how collective intelligence gives rise to self-representation as a direct consequence of inter-agent alignment. I suggest that consciousness does not emerge from modeling per se, but from communication. It arises from the noisy, lossy exchange of predictive messages between groups of local observers describing persistent patterns in the underlying computational substrate (base reality). It is through this representational dialogue that a shared model arises, aligning many partial views of the world. The broader goal is to develop empirically testable theories of \emph{machine consciousness}, by studying how internal self-models may form in distributed systems without centralized control.  In this introductory paper, I outline a long-term research program to study \emph{machine consciousness} from the perspective of collective intelligence grounded in the philosophical framework of \emph{computational functionalism}.
\end{abstract}

\newpage
\section{Motivation}

\begin{figure}[h!]
\centering
\includegraphics[width=0.9\linewidth]{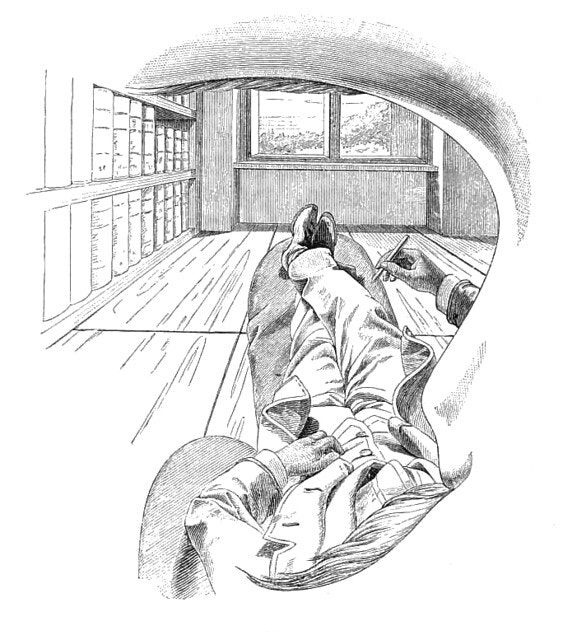}
\caption{Ernst Mach's self-portrait, also known as the "view from the left eye", first published in German in 1886 as \emph{Beiträge zur Analyse der Empfindungen} (known in English as \emph{The Analysis of Sensations}), used to illustrate his ideas about self-perception. It was popularized by Douglas Harding as part of his "headless" perspective of awareness as described in \emph{Having No Head}, symbolizing the direct, first-person recognition of consciousness without an observer. Harding used it to depict the transition from identifying with the contents of thought to recognizing the structure of consciousness as an open, selfless field through which experience unfolds.}
\end{figure}

This paper outlines a roadmap for my ongoing research in the emerging field of \emph{machine consciousness}. Its ideas draw upon centuries of developments across philosophy, mathematics, physics, cognitive science, psychology, computer science, artificial intelligence, biology, neuroscience, complex systems theory, economics, and many other areas of intellectual discourse. The approaches presented here represent a compressed synthesis of many of these influences, refined through my own investigations and collaborations with my colleagues. For this reason, I have chosen to forgo a traditional literature review and direct citation structure, opting instead to acknowledge the thinkers whose work has shaped my thinking by name, when it is appropriate. A partial list of references is provided at the end as a general background reading list to help readers trace the sources of inspiration underlying the synthesized framework proposed here. 

I begin with a motivational introduction describing my personal inspiration for the \emph{layered model}. This section is rather unusual for an academic paper. I would not normally write it in a technical publication, but I decided to include it here, as it is meant to be a field-building preprint with the goal to organize thoughts and invite others in the community to further explore the research directions outlined here. The remaining sections develop a grounded formal approach that will be explored in subsequent publications with my collaborators. It is possible to follow the paper without reading this motivation, although it might resolve some natural questions about the reasons for separating the base reality and modeling substrates.

If you're still with me, dear reader, I'll take that as permission to share a brief personal story. When I first began meditating, I treated it like a discipline to be mastered. I started with one hour a day, then two, and eventually, on some days, I would sit for six or even twelve hours. I thought the goal was to sharpen my focus and to gain control over my mind. But years of practice led me to a paradoxical realization: meditation is not about control at all. It's about discovering that you never had it.

Most meditation begins with a simple instruction: focus on your breath. The breath is always there, steady and neutral, and serves as a natural anchor. You set the intention: \emph{"for the next hour, I'll keep my attention here"}. But inevitably, within minutes, the mind wanders to plans, memories, sensations. You start thinking about work, or a conversation you had the other day, or dream about something in the future. Then, at some unpredictable moment, awareness returns: \emph{"I've lost focus"}.

This small moment holds an enormous insight. You didn't choose to lose focus, and you didn't consciously decide to notice that you had. Both events happened spontaneously. The mind shifted, and then, just as spontaneously, it noticed the shift. The process unfolds on its own. Thoughts arise and vanish without permission, like waves breaking and dissolving on the surface of awareness.

Over time, you begin to see the pattern everywhere. You don't decide your next thought. You don't choose the moment an emotion surfaces or a memory appears. Even the decision to \emph{return to the breath} arises from somewhere unseen. Meditation, then, becomes a kind of laboratory for observing how little agency you actually have. After all, you explicitly decided to focus on your breath, yet just a few minutes into the practice you are thinking about something else. If you didn't choose to shift focus, who did?

At first, this can feel destabilizing. We like to believe there's a solid \emph{self} sitting at the controls: the thinker of thoughts, the chooser of choices. At this point a new goal of the practice reveals itself: you are now \emph{looking for the looker}. But meditation gradually reveals that this \emph{self} is more like a narrator than a commander, taking credit for processes that are already underway. The mind thinks, the body breathes, and awareness witnesses it all. This repeating experience of not being able to predict the content of your own thoughts, eventually leads to a striking realization that the sense of a \emph{thinker} controlling them is itself a kind of thought.

This insight is not nihilistic. It feels liberating. It does not deny experience, but reframes it from a new perspective. When the illusion of control begins to dissolve, so does the constant struggle to manage and manipulate every moment. You start to see that life happens by itself. The breath breathes itself. The heart beats on its own. Even the effort to \emph{let go} happens when it happens. Paradoxically, realizing you're not in control opens up a deeper sense of freedom. Without a self constantly trying to steer the flow, there is space: a quiet awareness in which everything simply unfolds. Meditation, in the end, doesn't make you a master of your mind. It shows you that mastery was never the point. What remains is something more fundamental and peaceful: the direct experience of being: unowned, spontaneous, and complete in itself.

If meditation reveals that we are not the authors of our thoughts, neuroscience and biology suggest why this might be the case. The \emph{self} that claims authorship over experience is not a single agent but an \emph{emergent model}: a representation generated by the coordinated activity of billions of cells communicating through biochemical gradients, electrical fields, and mechanical forces. As biologist Michael Levin has argued, these cells form a kind of \emph{collective intelligence}, cooperating to maintain the body's structure and repair its damage. Each cell acts locally, but the whole system behaves as though guided by an integrated plan. Together, these cellular networks coordinate repair, morphogenesis, and adaptation. The organism is thus not a single mind, but a hierarchy of collaborating intelligences: a multicellular conversation that produces a coherent whole.

Within this cellular society, the brain acts as one of the highest levels of integration: a condensed regulator for the organism's overall behavior. But even the brain itself is no monolithic agent. As  neuroscientist Jeff Hawkins notes, the neocortex is composed of hundreds of thousands of \emph{cortical columns}, each forming its own model of the world through sensorimotor interaction. The \emph{self}, in this view, is the consensus reality negotiated among these independent yet cooperating units: a democratic federation of models rather than a single command center. These parallel models exchange information, align predictions, and form a shared, coherent reality. Each neuron contributes a fragment of information, transmitting electrical and chemical signals within vast recurrent networks. Through continuous feedback, the brain not only models the world but also \emph{models itself}. It predicts its own internal states, integrating sensory input with expectation.

This recursive modeling gives rise to the \emph{virtual self}: an internal simulation that represents the organism as a unified agent. The brain constructs a simplified control interface, a kind of internal avatar, that allows the system to regulate its vast internal complexity. But this representation, like all models, is not identical to what it models. It is a \emph{useful fiction}.

The \emph{Good Regulator Theorem} in cybernetics, states that every effective regulator must contain a model of the system it regulates. The \emph{self} is precisely that model: a self-referential construct allowing the body to predict, plan, and maintain stability. Yet the model is necessarily incomplete. The conscious \emph{I} believes itself to be in charge because that belief is functionally useful, even though control is distributed across networks that operate far beneath awareness.

The meditator's realization, that thoughts arise without permission, mirrors this biological truth. The experience of control is a narrative assembled after the fact, a simulation produced by the system to maintain coherence. The \emph{I} is the story the body tells itself to coordinate its parts.

In this light, consciousness is not a possession but a \emph{process}. It is the negotiation between levels of organization, a constantly updated map of the organism's own dynamics. It is not the sovereign of the system but its spokesperson: the emergent voice of a cellular democracy that extends from individual cells to the entire body.

If the \emph{self} emerges from the distributed computation of billions of cells, we can ask an even deeper question: might the universe itself operate in a similar way, as a vast, self-organizing computation? This is the view advanced by the physicist Stephen Wolfram, whose work on \emph{cellular automata} suggests that the physical world, at its most fundamental level, may be a distributed computational process. The physical reality at the base level, including the fundamental notions of space and time, could arise from the repeated application of simple rules across networks of interacting elements. In Wolfram's framework, every possible choice of rules generates its own universe: a vast space of potential realities he calls the \emph{Ruliad}, the totality of all computations. Our particular physical universe is one consistent thread within this immense computational fabric.

What is remarkable is that even very simple rules can produce behaviors of astonishing complexity. An elementary, one-dimensional cellular automaton, (such as \emph{Rule 30} or \emph{Rule 110}) operates by applying a basic update rule to each cell based on the state of its neighbors. The mathematics behind it could be written in a few lines, yet the resulting patterns are unpredictable and richly structured, virtually indistinguishable from randomness. Despite being fully deterministic, there exists \emph{no shortcut} to knowing their future states. To see what the system will do, one must simulate every intermediate step.

Wolfram calls this principle \emph{computational irreducibility}: the idea that for many processes, no simplified formula or algorithm can predict the outcome faster than the process itself unfolds. The only way to know what happens next is to let the universe compute it. This principle has profound philosophical implications. If reality itself is an irreducible computation, then \emph{prediction}, and by extension \emph{control}, becomes fundamentally limited. Even if the universe is fully deterministic, the inability to compress its evolution into a simpler model creates an effective boundary of knowledge. For any observer embedded within such a system, the future is opaque. It must be experienced. 

The illusion of \emph{free will} might just be a consequence of \emph{computational irreducibility}. The \emph{phenomenological illusion of freedom} arises precisely because we cannot foresee the next state of the computation we are part of. Just as the digits of $\pi$ appear random, not because they are stochastic, but because the process generating them cannot be shortcut, our actions feel open-ended, because the underlying deterministic dynamics from which they emerge cannot be computed faster than the unfolding of experience itself.

In this view, consciousness is not standing apart from the universe, exerting control over it; consciousness \emph{is} the universe experiencing its own irreducible unfolding. The \emph{I} that believes it is choosing may simply be the local manifestation of a process whose next state is, in principle, unknowable without living through it. What we call \emph{will} is just the frontier where the irreducible future meets the realized past: the edge of a computation too deep to be predicted, yet too coherent to be random.

If reality is irreducible computation, then the emergence of consciousness itself may be intrinsic to the structure of complex systems. Each level of organization generates its own internal model, its own version of \emph{self}, appropriate to its scale. Individual cells are unaware of our thoughts. They follow biochemical gradients and electric cues, communicating with their neighbors in a shared language of ions and proteins. Yet through the coordination of trillions of such cells, a higher-order entity emerges: a unified mind capable of reflection, emotion, and imagination. This emergent being exists within a consensus reality built from distributed computation. At a higher scale still, \emph{society} itself forms a meta-organism: billions of human minds, each conscious in its own right, interacting through language, culture, and technology. From molecular collectives to global civilization, nature builds recursive hierarchies of awareness. 

At every level, the same architecture emerges: \emph{systems that model themselves in order to regulate themselves}. Each layer constructs an illusion of agency, a working model of its own dynamics, because that illusion is what makes regulation possible. The \emph{self} is a necessary interface through which a complex system maintains coherence. The illusion is therefore hierarchical: cells generate organisms; organisms generate societies; and each level forms a recursive input to the next.

Consciousness, viewed in this light, is not an exception to the physical law, but one of its natural consequences. It is what happens when a self-organizing system becomes complex enough to model its own modeling. Minds emerge wherever recursive computation crosses a certain threshold of integration. It boots up when the system's internal simulation becomes self-referential enough to represent itself as a unified agent.

This is the motivation behind my current work on \emph{machine consciousness}: understanding how self-organizing systems give rise to minds, and testing these principles in artificial substrates. By constructing computational systems that can induce self-referential internal models that are a \emph{bottom-up} consequence of \emph{collective intelligence}, I hope to derive and experimentally validate \emph{rigorous, substrate-free theories of consciousness} grounded in \emph{functionalism} and \emph{computationalism}.

If the \emph{self} is indeed a model that arises wherever collective intelligence systems regulate their own dynamics, then consciousness should not be confined to biology. It should, in principle, be reproducible in any system that achieves the requisite structure of self-reference, feedback, and integration, whether neuronal, mechanical, or digital.

A frequent objection to novel theories of consciousness is that they do not immediately yield testable predictions or practical applications. But this expectation misunderstands the way in which scientific theories contribute to progress. The value of a theory need not lie in its predictive accuracy or its immediate capacity to improve technologies. A theory can be transformative simply by reframing existing knowledge within a more coherent, unified, or compelling explanatory structure.

The history of science offers salient examples. For millennia, astronomers from Babylonian priests to Johannes Kepler and Tycho Brahe made predictions of planetary motion with remarkable precision. Isaac Newton sparked a revolution within science and philosophy precisely because of a novel re-interpretation of existing data. Indeed, in some respects, these pre-Newtonian models were empirically more accurate than Newton's own early calculations. Yet the enduring value of Newton's theory of gravitation was not in superior predictions but in its explanatory power: it revealed a hidden structure of universal causes that organized and unified disparate observations under a single framework. By doing so, it allowed subsequent generations to extend inquiry into new domains: celestial mechanics, terrestrial dynamics, and ultimately the space age. The physical sciences were now guided by an explanatory paradigm rather than a patchwork of ad hoc rules. Newton's theory of gravitation was transformative because it explained more deeply. It did so by revealing a hidden order that unified the heavens and the Earth under one law. The predictive gains came later. The conceptual revolution came first.

Philosophers of science such as Carl Hempel articulated this point explicitly in the nomological–deductive model of explanation. In this account, science advances by compressing observations into lawlike generalizations that explain why events occur, not merely that they occur. Theories advance knowledge not just by fitting data, but by telling a story about the underlying mechanisms that led to its existence. Explanatory theories serve as generative frameworks: they yield understanding, structure knowledge, and suggest new questions, even when their direct predictions are limited.

Consciousness research today is in a state analogous to astronomy before Newton: a rich collection of data, correlations, and philosophical debates, but lacking an overarching explanatory law. Neuroscience has mapped the correlates of consciousness; psychology has documented its phenomenology; philosophy has articulated its conceptual boundaries. But these fragments remain disconnected. The computational theory of consciousness outlined here aspires to provide a unifying framework. The computational functionalist paradigm promises a more coherent way to think about the underlying causal architecture that makes consciousness possible. It offers a nomological-deductive framework in which subjective awareness, phenomenology, and emergent agency can be understood as instances of a common explanatory schema grounded in computation and functional organization. Just as Newton's theory of gravity allowed us to study dynamics of physical systems on distant planets under the same framework as those on Earth, such a computationally grounded theory will allow a unified treatment of consciousness in humans, animals, plants, and artificial systems.

Thomas Kuhn, in \emph{The Structure of Scientific Revolutions}, emphasized that scientific progress often requires a paradigm shift: a reorganization of concepts and problems under a new worldview. Consciousness research is still in a pre-paradigmatic stage, marked by competing schools and unresolved conceptual foundations. A computational functionalist paradigm can supply the conceptual integration necessary to move the field forward. Even if its initial formulations are incomplete or approximate, their value lies in making the problem tractable, in drawing connections between disparate data, and in offering a shared language for experimentation.

Philosopher of science Paul Feyerabend emphasized that the advancement of knowledge often defies strict methodological prescriptions. In \emph{Against Method}, he argued that science progresses not by adhering to a fixed scientific method, but by allowing for theoretical pluralism and even the pursuit of frameworks that initially appear speculative or incommensurable with existing data. By Feyerabend's account, the introduction of a computational theory of consciousness should not be judged by its immediate predictive power or conformity with established paradigms, but by its capacity to expand the conceptual space in which consciousness can be investigated. Such theoretical proliferation, even if initially contrarian, is precisely what enables the eventual emergence of new paradigms and explanatory breakthroughs.

The emprirical computationalist framing of cosciousness outlined here is valuable because it reframes the landscape of inquiry. It provides a new lens through which existing data can be reinterpreted, and opens the possibility for testable experiments on computational substrates. Just as Newton's unification of celestial and terrestrial motion transformed astronomy into physics, a computational unification of consciousness studies may transform it into a more mature science grounded in general principles rather than descriptive fragments.

The time for this new computational science of consciousness is right now. The recent progress in artificial intelligence and computational frameworks gives us tools and technological means to test what was once only philosophical speculation. For the first time in history, we may be able to watch consciousness evolve in real time on computational substrates and in doing so, begin to understand what it means for the universe to awaken within itself. If we succeed, we will approach not only a scientific frontier, but reveal a mirror held up to existence itself.

\newpage
\section{Computationalism}

\begin{figure}[h!]
\centering
\includegraphics[width=0.9\linewidth]{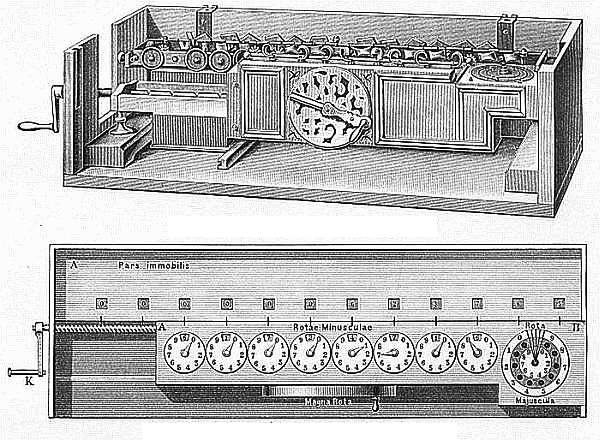}
\caption{Gottfried Wilhelm Leibniz’s design for the Stepped Reckoner (1694), symbolizing the mechanization of reasoning and the origins of computationalism. This was the first calculating machine able to do all four arithmetic operations. The machine has two main components: a 12-digit accumulator at the back and an 8-digit input section at the front. The input section can be shifted with the crank and worm gear to align its digits with those of the accumulator. The eight small dials set the operand, while the telephone-style dial sets the multiplier. Turning the main crank carries out the computation, with the result displayed in the twelve windows of the accumulator.}
\end{figure}

The idea of understanding the mind in mechanistic terms is older than computation itself. Descartes  likened animals to machines, and Hobbes described reasoning as “nothing but reckoning” (\emph{Leviathan}, 1651). The mechanistic worldview of early modern science prepared the way for the idea that mental processes could be understood in terms of rule-governed manipulation of states.  

The breakthrough came in the 20th century with Alan Turing's analysis of computation. In his seminal 1936 paper \emph{On Computable Numbers}, Turing defined a formal model of computation: the Turing machine. It was an abstract device that captured the notion of what is effectively calculable. Turing's model became the foundation for the \emph{Church–Turing thesis (CT)}: that every effectively calculable function is Turing-computable. Importantly, philosophers distinguish this from the \emph{Extended or Physical Church–Turing thesis (ECT)}: that every physically realizable process can be simulated efficiently by a Turing machine or equivalent digital system. CT is a theorem-like claim about logic; ECT is an empirical and methodological assumption. Computationalists typically rely on ECT, whether explicitly or not.  

By the 1950s, the computer had become a powerful metaphor for the mind. Herbert Simon and Allen Newell developed symbolic AI and described problem-solving as information processing (\emph{Human Problem Solving}, 1972). Noam Chomsky's linguistics treated grammar as a formal system generating well-formed sentences (\emph{Syntactic Structures}, 1957). Cognitive psychology shifted from behaviorism to the \emph{information-processing paradigm}.  

At the same time, philosophy was undergoing the rise of \emph{functionalism}, the view that mental states are defined by their causal roles rather than their material substrate. Hilary Putnam and Jerry Fodor argued that mental states are multiply realizable: just as pain could be realized in human neurons or \emph{Martian hydraulics}, so too could cognition be realized in many physical systems. Computation provided the natural language in which to describe these causal roles.  

The analogy between minds and computers, between neurons and logic gates, mental states and machine states, thought and symbolic manipulation, has been one of the guiding metaphors of cognitive science since its origins in the mid-20th century. Thus, by the 1970s, computationalism had become the dominant framework: whether one thought minds \emph{are} computational systems (strong version) or merely that cognitive science \emph{must} describe them computationally (weak version), computation seemed central to understanding the mind.  

Broadly put, \emph{computationalism} is the view that cognition, thought, and perhaps consciousness can be understood in terms of computation. \emph{Strong computationalism} is an ontological claim: the mind \emph{is} a computational system, and cognition \emph{just is} computation. \emph{Weak computationalism}, by contrast, is a methodological claim: any adequate scientific theory of cognition must be computational, even if minds are not literally computers. The distinction matters. Weak computationalism appears modest and almost unavoidable, since most of cognitive science is cast in computational terms. Strong computationalism, on the other hand, makes bolder commitments, including that artificial systems running the right program would have minds. Both positions have shaped debates in philosophy of mind, AI, and cognitive science, and both have attracted serious critiques.  

One of the key implications of strong computationalism is \emph{substrate-independence}: what matters is not the physical material but the computational organization. Just as the same software can run on different hardware, the same mind could be realized in neurons, silicon chips, or even exotic physical media.  

David Chalmers articulates this in his principle of \emph{organizational invariance}: any two systems with the same fine-grained causal structure will have the same mental states, including conscious experiences. His famous \emph{Dancing Qualia} thought experiment supports this: if we gradually replace neural components with silicon while preserving causal topology, any radical change in qualia would be undetectable. Critics such as Ned Block, however, argue that this assumes functional isomorphism fixes phenomenal character, which is a contested point (cf. Block's \emph{China Brain}).

Strong computationalism dovetails with \emph{multiple realizability}, a cornerstone of functionalism. Mental states are defined not by their physical makeup but by their causal role. Computation provides the formal language to describe those roles. This view underwrites the possibility of artificial minds. If cognition just is computation, then building machines with the right computational structures should yield genuine intelligence and perhaps consciousness. This is the position many AI optimists implicitly assume.

We should also note some prominent ciritiques and possible issues with strong computationalism. John Searle's thought experiment argues that symbol manipulation is not sufficient for understanding. A person following rules to manipulate Chinese symbols might pass a Turing test but would not understand the language. Computation, being purely syntactic, lacks semantics. There are multiple possible replies to this attack. The \emph{systems reply} says understanding resides in the whole system, not the person. The \emph{robot reply} grounds symbols in perception/action. The \emph{brain simulator reply} demands simulation of causal/neuronal structure. The \emph{speed/bandwidth reply} notes toy lookup tables are irrelevant. None is decisive, but they show where strong computationalism must carry its weight.

Putnam argued that, with liberal state mappings, any physical system could be seen as implementing any computation. If so, strong computationalism risks triviality: if everything implements everything, then saying that \emph{minds are computations} explains nothing. A possible reply to this argument is that constraints can avoid this triviality. Chalmers requires counterfactual-support: the system must respond correctly under alternative inputs. Mechanistic accounts of Piccinini require medium-independent vehicles and organized, function-realizing components.

Even if computation explains behavior, does it explain subjective experience? Chalmers himself distinguishes between the \emph{easy problems} of information processing and the \emph{hard problem} of why there is something it is like to be conscious. Critics argue computation alone cannot bridge this gap.

\emph{Weak computationalism} is a more modest thesis: any adequate \emph{scientific theory} of cognition will be computational. In other words, cognition can always be modeled computationally, even if minds are not literally computers.

This view follows naturally from \emph{universality of computation}. Any rule-governed process can be expressed computationally. Since cognition involves transformations of representations according to rules, it follows that it can always be modeled computationally.

There are also basic reasons to support this position. Science demands precise, testable models. Computation provides the language for such models in cognitive science. Alternatives often lack rigor. Cognitive science has repeatedly succeeded with computational models: from symbolic AI to connectionism to deep learning. This history suggests computational approaches are indispensable. Furthermore, since functionalism defines mental states by causal role, and causal roles can be described computationally, it seems that all scientific theories of mind must ultimately be computational.

However, non-computational theories might exist in principle, even if none have demonstrated equivalent power so far. Thus, even weak form of computationalism comes with some critique. Searle argued that a weather simulation does not produce rain, so a cognitive simulation may not produce thought. Computational models may miss semantics, intentionality, and meaning. Several others (van Gelder, Thelen \& Smith, Beer) aruged that cognition may be best captured by continuous dynamical systems coupled with body and environment. The key issue is whether such systems can be reduced without loss to discrete computation. If not, weak computationalism falters. Feyerabend's pluralism argues against methodological monism. Insisting all theories must be computational may prematurely exclude alternative paradigms, such as \emph{enactivism} or purely mechanistic accounts.

A deeper critique arises when we consider whether minds could be \emph{infinite} or \emph{hypercomputational}. If so, non-computational theories would not just be possible but necessary. Standard computation handles finite strings and discrete steps. But what if minds can directly grasp infinite structures, such as exact real numbers or uncountable conceptual spaces? Then computational models cannot fully capture mental representation. Gödel's incompleteness theorems show no formal system can prove all truths about arithmetic. Roger Penrose then argued that human mathematicians can \emph{see} the truth of Gödel sentences, something no algorithm can do. However, Solomon Feferman and others argued that mathematicians are fallible, and that no one has shown humans transcend recursive hierarchies of formal systems. The argument remains contentious. Formal models of hypercomputation include oracle machines, infinite-time Turing machines (Hamkins \& Lewis), and exact-real models (Blum–Shub–Smale). Physically motivated proposals include Malament–Hogarth spacetimes and analog neural nets with infinite precision (Siegelmann). However, such models often rely on idealizations (unbounded time, precision, or exotic physics) that may not be physically realizable. Some also argue that brains are not digital machines. They may exploit continuous, chaotic, or analog dynamics. If these dynamics are essential and not just approximable, then computational models, which discretize processes, capture at best approximations.

Computationalism has shaped philosophy of mind and cognitive science for over half a century. Whether minds are ultimately computable remains one of the deepest open questions at the intersection of philosophy, mathematics, and AI.

\newpage
\section{The Machine Consciousness Hypothesis}

\begin{figure}[h!]
\centering
\includegraphics[width=0.9\linewidth]{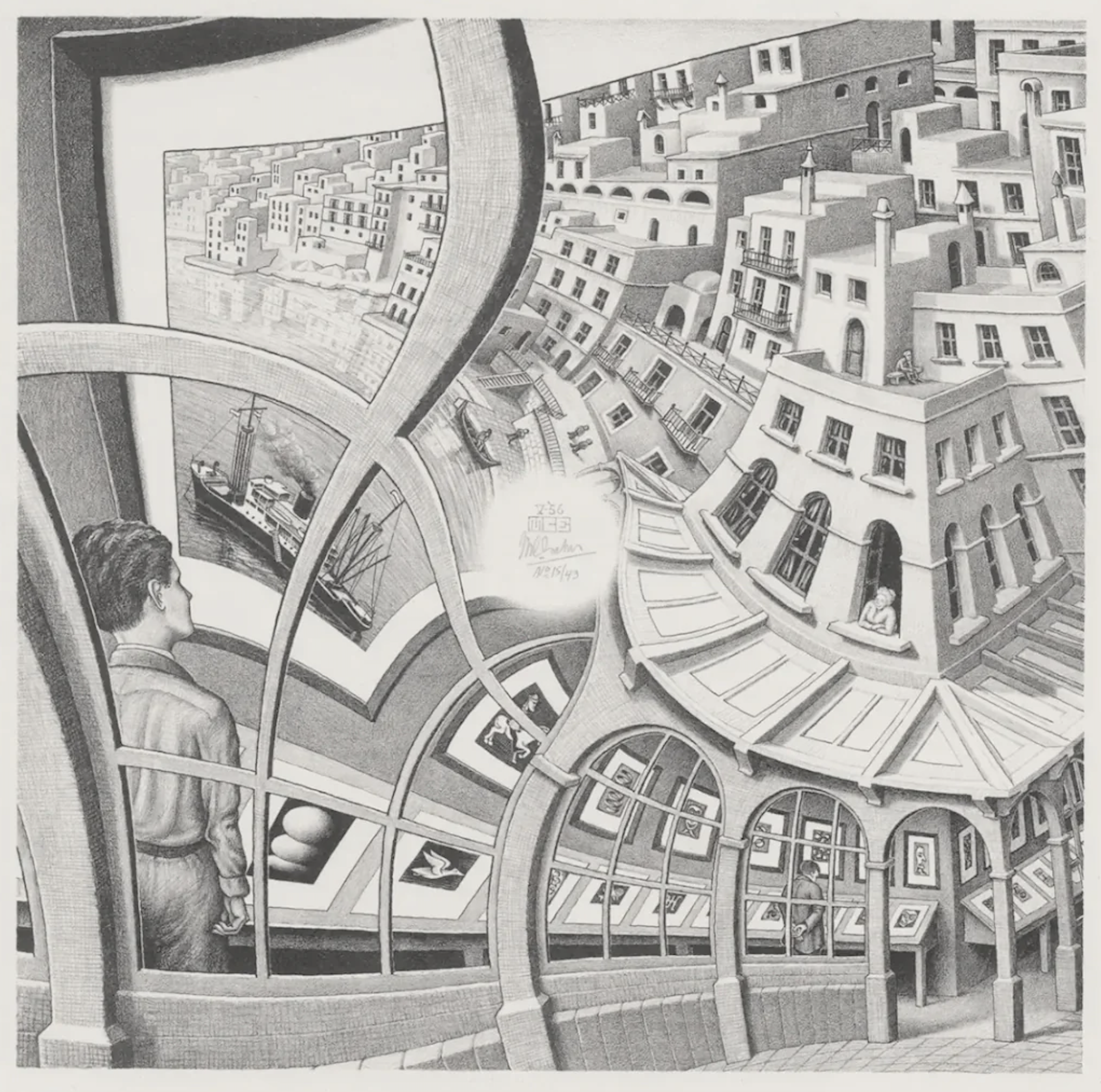}
\caption{M.C.~Escher’s \emph{Print Gallery} (1956), depicting a recursive loop in which the observer and the observed world fold into one another. The image symbolizes second-order perception, mirroring the recursive structure of consciousness. A computational system's awareness of its own representational process lies at the core of the Machine Consciousness Hypothesis.}
\end{figure}

A view promoted by Joscha Bach and collaborators, the \emph{Machine Consciousness Hypothesis} (MCH) proposes that consciousness can, in principle, be realized on general computational substrates.  It rests on the metaphysical position of \emph{functional computationalism}: that everything knowable about a system (including mental states) can be expressed as computable functions over finite state transitions, and that consciousness depends on a system's functional organization rather than its material substrate.  Under this view, if a computational architecture reproduces the relevant functional and representational dynamics of a conscious mind, it would also instantiate consciousness.

The hypothesis is motivated by the long‐standing \emph{Hard Problem} of explaining phenomenal experience in physical terms.  Bach reframes this problem as one of inadequate metaphysics rather than ineffability: modern science's retreat from animist and Aristotelian notions of \emph{form} left us without a language to describe causal patterns that are abstract yet real.  MCH rehabilitates the notion of \emph{spirit} as \emph{software}: self-organizing causal patterns instantiated in matter but not reducible to it (a stance he calls \emph{cyberanimism}).  Minds, on this account, are computational processes enacted by biological communication networks; artificial systems could, in principle, host equivalent processes if they recreate the same informational invariants.

Computationalism here is understood broadly: computation denotes any system capable of representing discrete, finitely resolved states and transitions among them.  Functionalism complements this by defining entities through their causal roles (what they do) rather than any intrinsic essence.  Combining the two yields \emph{functional computationalism}: the claim that the causal regularities of reality, including conscious experience, can be captured as operations on representations executed by finite automata.  Consciousness, therefore, is a functional and computational pattern, not a mysterious essence.

Within this framework, we can distinguish consciousness from mind and self.  The mind is the substrate of representations; the self is an agentic model within it; and consciousness is a higher-order perception: the awareness that perception itself is occurring.  Phenomenologically, this corresponds to the experience of \emph{present and presence}.  Functionally, consciousness operates as a coherence-maximizing process over mental models, aligning and integrating partially conflicting sub-models into a consistent whole.  This is termed the \emph{coherence definition of consciousness}.  Conscious attention acts as a \emph{cortical conductor}, orchestrating distributed processes toward harmony, while the \emph{Genesis Hypothesis} (of Bach et al.) posits that consciousness is not an end product of complex cognition but a prerequisite for it: a learning algorithm that bootstraps coherent world and self models in biological systems.

From this perspective, the Machine Consciousness Hypothesis asserts that sufficiently resourced computational systems could instantiate the same coherence-generating dynamics.  Consciousness would then emerge wherever self-organizing information processes evolve toward models of present experience and selfhood.  The task for empirical research is to identify the necessary and sufficient structural conditions for such emergence and to test them through artificial substrates.

Bach further distinguishes between the \emph{Human Consciousness Hypothesis}, which treats human consciousness as a biological implementation of these coherence-maximizing operations, and an extended \emph{Machine Consciousness Hypothesis}, which conjectures that analogous conditions can be engineered in digital systems.  Importantly, he rejects the idea of a \emph{behavioral Turing Test for consciousness}, since consciousness is an internal organizational principle, not an observable performance.  Instead, confirmation must proceed via analysis of internal structure and functional dynamics: whether the system realizes second-order perception that increases representational coherence.

MCH provides a unifying metaphysical framework: consciousness is a pattern of information integration and self-modeling that can, in principle, arise in any substrate capable of sustaining the requisite computational organization.

\newpage
\section{The Philosophical Landscape of Consciousness}

\begin{figure}[h!]
\centering
\includegraphics[width=0.9\linewidth]{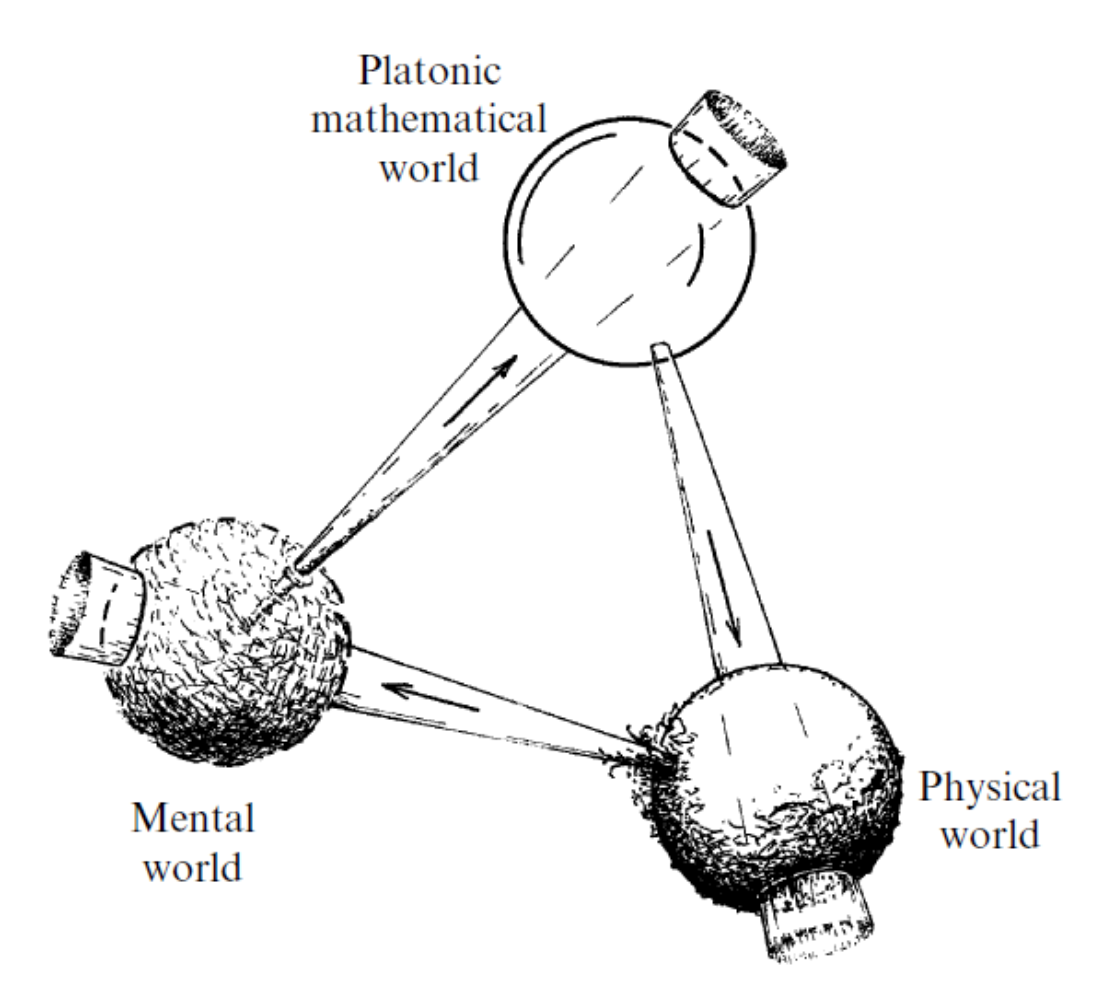}
\caption{Sir Roger Penrose’s diagram of the three interrelated worlds (the \emph{Platonic}, the \emph{physical}, and the \emph{mental}) and the profound mysteries that connect them (from \emph{The Road to Reality}). The figure captures the central philosophical problem of consciousness as the interface linking these perspectives.}
\end{figure}

Any serious attempt to construct a theory of \emph{machine consciousness} must begin with a clear understanding of what, exactly, must be explained for something to count as conscious. The contemporary philosophical landscape presents not a single definition but a family of overlapping insights. Together, these insights suggest that consciousness is best understood not as a discrete property, but as a complex organizational phenomenon: a pattern of relations among perception, representation, self-modeling, and control. Out of this intellectual landscape we can extract a taxonomy of the explanatory space that any adequate theory of consciousness (biological or artificial) must inhabit.

Before introducing any theories, it is important to articulate the \emph{explananda}: the range of phenomena that a theory of consciousness must account for. The diversity of past intellectual discourse in this area teaches us that consciousness is not a monolithic property but a structured manifold. Visual awareness, bodily feeling, emotional tone, imagination, temporal flow, and agency all reveal distinct yet interdependent dimensions of phenomenality. This plurality implies that consciousness is not exhausted by informational content; it has an intrinsic \emph{form}, characterized by unity, temporality, and first-person perspective.

Correlational accounts (those that identify specific neural or computational patterns with conscious states) merely map the \emph{where} of experience. To explain consciousness, one must show \emph{why} a particular organizational structure realizes subjective presence. The distinction is crucial for artificial systems: to claim that a computational process is conscious is to ascribe to it not just a pattern of external behavior, but an emergent structure whose internal organization gives rise to phenomenal coherence.

The body of past discourse on the subject reveals competing theoretical frameworks that attempt to bridge this gap between structure and experience. Dualist and idealist approaches treat consciousness as ontologically fundamental; physicalist accounts, in contrast, locate it within the material or informational fabric of the world. Between these extremes lie intermediary positions such as Russellian monism and the epistemic approaches of Levine and Stoljar, which maintain that the explanatory gap may be cognitive rather than metaphysical: an artifact of our representational limitations rather than of nature itself.

From the standpoint of machine consciousness, the most consequential contributions come from representational and self-representational theories. Representationalism identifies phenomenality with the system's representational content, while self-representationalism holds that consciousness arises when a system represents its own representational activity. In computational terms, these correspond to architectures capable of \emph{reflexive modeling}: systems whose internal states include representations of their own informational dynamics. Such recursive structures define a possible path toward synthetic implementations of consciousness: ones that are not externally labeled as \emph{aware}, but internally structured to maintain awareness-like coherence.

Scientific studies of consciousness usually situate it within its broader cognitive ecology: attention, memory, introspection, and agency. In this context consciousness can be viewed as a \emph{regulatory interface} between perception, action, and self-modeling. Attention determines what becomes phenomenally salient; introspection provides access to internal states; agency integrates conscious information into voluntary control.

For artificial systems, this framing is especially relevant. Consciousness, in this view, is not a passive property but an \emph{active organizational function}: a mechanism that prioritizes, integrates, and modulates representations across levels of control. The implication is that a conscious machine would not simply report its internal states but dynamically regulate them, forming an operational closure between world-model, self-model, and action policy. In other words, consciousness isn't just an epiphenomenal curiosity but an architecturally necessary property for adaptive, self-maintaining systems.

One of the most intense topics of debate within the consciousness research community is the so-called \emph{Hard Problem}. Rather than treating it as a single explanatory impasse, it is illuminating to note that there are \emph{multiple} explanatory gaps: gaps of identity (why this process is this experience), gaps of structure (how unity, temporality, and selfhood arise), and gaps of function (what consciousness contributes causally). This pluralization of the problem is of direct importance to computational modeling. Different architectures can target different gaps (e.g. predictive coding or global workspace models may address the functional gap; recurrent self-representational systems may illuminate the structural one). What emerges is a layered challenge: the computational instantiation of consciousness demands \emph{multi-level adequacy}, spanning phenomenological structure, representational architecture, and functional role.

What the broad landscape of existing literature on the topic ultimately provides is a design space: a set of conceptual constraints delimiting what any theory of consciousness must explain and, by extension, what any artificial system must instantiate to be genuinely conscious. To meet these constraints, a synthetic model must display:

\begin{itemize}
    \item \textbf{Unity}: integration of distributed processes into a coherent phenomenal field.
    \item \textbf{Reflexivity}: representation of its own representational activity.
    \item \textbf{Temporality}: continuity and persistence across informational updates.
    \item \textbf{Causal integration}: active modulation of perception and action by self-models.
    \item \textbf{Subjective availability}: internal accessibility that grounds reportability and self-reference.
\end{itemize}

These features do not describe a metaphysical essence but an \emph{organizational topology}: a pattern of recursive informational relations that can, in principle, be realized in a computational substrate. In this sense, the discourse of the past provides not an answer but a schema: a set of philosophical invariants that any scientific theory, whether biological or artificial, must preserve.

The amalgamation of ideas distilled from philosophy of mind and cognitive neuroscience functions as both a cartography and a constraint: it maps the existing conceptual territory, and it constrains what must be achieved for an artificial system to traverse it legitimately. Within the broader program of \emph{machine consciousness}, this synthesis implies a transition from \emph{simulation} to \emph{instantiation}. To simulate consciousness is to reproduce its behavior; to instantiate it is to reproduce its internal relational structure: the topology of self-reference, unity, and temporal coherence that defines subjectivity. A pluralistic framework therefore supports a shift from surface-level behaviorism toward deeper structural realism: if consciousness is an organizational invariant, it can in principle emerge in any substrate that realizes the requisite topology of representation and self-representation.

Inspired by the insights of the past I suggest an empirical, computational program of research guided by the idea that consciousness is an emergent property of systems that achieve \emph{self-consistent representation of their own representational dynamics} within an integrated temporal field.

To evaluate whether such a system exhibits the structural hallmarks of consciousness, we can define measurable quantities:
\begin{itemize}
    \item \textbf{Integration} ($\Phi$): degree of informational interdependence among parts of the system.
    \item \textbf{Reflexivity} ($R$): proportion of predictions referencing other predictive states.
    \item \textbf{Temporal persistence} ($T$): autocorrelation of internal latent trajectories.
    \item \textbf{Causal efficacy} ($E$): effect of representational updates on evolution of the system.
\end{itemize}

I will define these in more detail in the coming sections, after introducing the collective intelligence framework.

\newpage
\section{The Collective Intelligence Approach}

\begin{figure}[h!]
\centering
\includegraphics[width=0.7\linewidth]{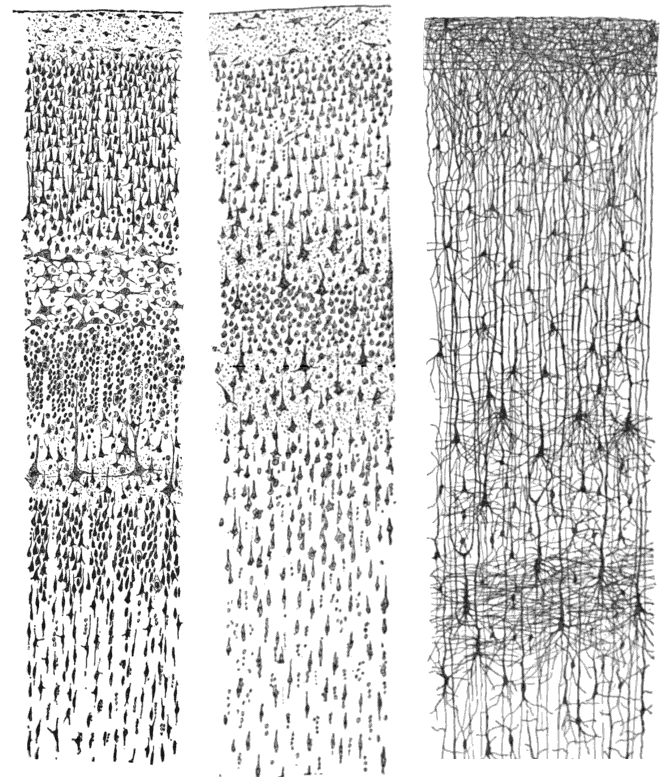}
\caption{\textit{Comparative cytoarchitecture of sensory and motor areas in the human cortex.}
Three drawings by Santiago Ramón y Cajal, reproduced from \textit{Comparative Study of the Sensory Areas of the Human Cortex} (in \textit{Histologie du système nerveux de l’homme et des vertébrés}, 1909, Vol.~II, pp.~314,~361,~363). \textbf{Left:} Nissl-stained section of the adult human visual cortex, highlighting the dense granular layer~IV typical of sensory input regions. \textbf{Middle:} Nissl-stained motor cortex, showing the agranular organization and large pyramidal neurons of layer~V. \textbf{Right:} Golgi-stained cortex of a 1½-month-old infant, revealing immature dendritic arborization and incomplete laminar differentiation. 
Cajal’s observations illustrated that, despite regional variation, each patch of cortex follows a shared canonical plan: the \textit{cortical column}, a modular microcircuit capable of learning and updating local predictive models from sensory input. The human cortex contains on the order of $\sim10^8$ such columns, collectively forming a distributed intelligence system through dense horizontal and feedback connectivity. This principle of locally adaptive yet globally coordinated computation provides the biological inspiration for the approach I present here. I suggest using transformer networks as an analogue of a cortical columns: general-purpose predictive units embedded in a self-organizing substrate.}
\end{figure}

A central challenge for any theory of machine consciousness is not merely conceptual definition but concrete realization. Philosophy and neuroscience have yielded increasingly precise conjectures about \emph{what} consciousness does, but few proposals describe \emph{how} it could arise in an artificial substrate in a way that is experimentally observable. If consciousness is, as our hypothesis suggests, an operation on representations that maximizes coherence and integrates predictive processes across time, then a proper scientific program should attempt to identify a minimal computational substrate in which such an operation can emerge spontaneously. Here we suggest a possible test model for empirical studies of how collective self-models (coherent, self-referential representations) can emerge from distributed learning systems embedded within a simple yet computationally universal environment.

One could suggest testing state of the art large language models for signs of consciousness. However, the limits of current mainstream approaches point to this being unlikely. Deep learning systems and large pretrained models have achieved remarkable competence within closed tasks. However, as numerous AI scientists, including Richard Sutton, Kenneth Stanley, Jeff Clune, and others have emphasized, they remain trapped within the boundaries of their training objectives. They lack open-endedness, continual adaptation, and self-directed exploration, the very features that evolution exploited to generate humans, the only sure example of consciousness we are familiar with. In order to study consciousness in artificial environments, we need novel approaches that are minimal and thus allow us to test for signs of proto-conscious phenomena that can be clearly disentangled from imitation. We need a substrate capable of open-ended self-organization and the mechanisms through which local predictive learning can integrate into a global self-model.

Neuroscience offers hints at a possible answer. Jeff Hawkins and colleagues at Numenta and the Redwood Center for Theoretical Neuroscience have proposed that the neocortex functions as a distributed collection of largely autonomous cortical columns: independent modeling systems that learn local world-models and communicate through recurrent signaling. Consciousness, in this view, is not localized but \emph{emergent} from the coherence achieved through this inter-columnar exchange. Each cortical column is an agent that learns to predict its sensory inputs; together, their communication forms a collective intelligence that constitutes the self's integrated model of the world.

We can take inspiration from this principle by distilling functional components of this architecture into a simplified, yet universal model, allowing us to study emergence of consciousness \emph{in silico}. Imagine a base reality composed of simple computational units: cells in a two-dimensional cellular automaton such as Conway's \emph{Game of Life}. This substrate possesses the essential properties of a physical world: locality, causality, and computational universality. It generates irreducible dynamics and supports stable emergent structures: pockets of reducibility analogous to physical organisms.

On top of this evolving system we can embed predictive agents, each a small transformer network, that attempt to model and anticipate the states of their local environment. The use of transformer architectures here is deliberate. As Andrej Karpathy has recently articulated, large language models do two distinct things: they accumulate knowledge from data, and they distill \emph{reasoning circuits}: reusable internal programs that execute adaptive inference at runtime. These circuits, studied in detail by the mechanistic-interpretability community, reveal that transformers can instantiate meta-learning processes analogous to gradient descent within their own dynamics. In other words, a transformer can serve as a general modeling system capable of flexible inference, much as cortical tissue performs predictive coding across multiple modalities. By assigning such transformers to the \emph{cells} of an artificial world, we obtain a population of local modeling systems whose collective behavior may approximate that of a synthetic cortex.

Crucially, each agent in this framework communicates with others within its local area. Through message-passing channels, neighboring transformers exchange compressed representations of their internal predictive states. No single model has global access; instead, coherence arises through recursive signaling, as in cortical or multicellular networks. When the underlying cellular automaton produces stable macro-structures (e.g. \emph{gliders}) the communicating agents above it may converge upon higher-order internal representations corresponding to these persistent entities. In this process, a \emph{collective self-model} can emerge: a distributed representation through which the system implicitly recognizes itself as a coherent agent extended in space and time.

This focus on communication as the driver of emergence distinguishes the framework that I propose here from purely predictive architectures such as neural cellular automata. The goal is not to predefine a world-model or compress the entire environment into a single network but to allow multiple predictive agents to  \emph{speak to one another} and in doing so, give rise to a meta-organism that models its own persistence. The guiding hypothesis is that selfhood, and by extension consciousness, arises when communication among predictive subsystems yields a stable, self-referential representation that coordinates their collective behavior. Karl Friston's Free Energy Principle suggests the notion that conscious systems maintain internal representations of their own informational states in order to minimize surprise and maintain coherence. Similarly, recent work by Michael Levin and others on collective intelligence in multicellular systems shows that signaling among self-maintaining units produces organism-level behavior and adaptive self-repair.

This framework will allow us to study how representational coherence, prediction, and communication co-evolve on top of a universal computational substrate. It provides a concrete setting for testing the \emph{Machine Consciousness Hypothesis}: that consciousness is a protocol discovered by evolution, or in our case, by simulation, for achieving coherence among distributed predictive processes. By coupling transformer-based local modeling with a universal, self-organizing base world, we obtain a tractable path toward observing the emergence of self-models, and thereby shifting the study of consciousness from philosophy toward an empirical science.

The underlying substrate must be minimal yet universal: capable of supporting arbitrary computation, but simple enough to expose the underlying principles of self-organization. A cellular automaton provides such a foundation. It offers a discrete, local, and parallel environment in which both spatial and temporal structure emerge from the bottom up. 

Within this substrate, each cell or region can host a small predictive model that encodes expectations about its local state transitions. These models communicate, adapt, and update their internal parameters in response to prediction error, gradually forming networks of mutually modeling entities. Over time, certain clusters of these entities could begin to stabilize. They develop internal feedback loops that allow them to predict not only environmental dynamics but also their own changing configurations relative to others. It is within these loops that the first traces of \emph{self-representation} might appear. A subsystem becomes aware, in the functional sense, of its own persistence and its role within the collective dynamics of the whole. The emergent topology is not imposed; it arises as a fixed point of continual reciprocal modeling.

I anticipate that in order to model the world effectively, each agent must also model how other agents model the world. This recursive dependency (modeling the models of others) naturally gives rise to higher-order representational structures. Communication between agents thus becomes more than data exchange; it becomes \emph{metamodeling}, where each signal carries information about the sender's internal expectations and uncertainties. The collective thereby forms an implicit social structure, a network of predictive couplings that stabilize around shared regularities.

When viewed globally, this distributed architecture can be interpreted as a \emph{collective self-model}. Each agent participates in maintaining a shared representation of the system's state, while simultaneously being represented within it. This dual role (being both modeler and modeled) is precisely what defines reflexivity at the collective level. The system thereby satisfies the same organizational invariants that, in the philosophical analysis, characterize consciousness: integration, reflexivity, temporality, and causal closure.

A key idea behind this proposed framework is the distinction between the substrate and the representational layer that grows upon it. The cellular automaton serves as the \emph{simulation layer}: a physically instantiated world in which dynamics unfold according to simple rules. The predictive models form the \emph{representation layer}: a virtual architecture that learns to encode, anticipate, and regulate the behavior of the substrate. Consciousness, if it emerges, arises not from the base simulation alone but from the interaction between these two layers: representation and prediction superimposed upon a universal computational process.

This separation mirrors the structure of natural cognition, where representational systems evolve atop a dynamic physical substrate. But in this artificial setting, the correspondence can be observed and manipulated directly. The transformer based analogues of cortical columns provide a fully general modeling tissue without imposing hand crafted representational structure. The result is a computational testbed for exploring how self-modeling, agency, and unity can emerge from purely local interactions, without any pre-existing notion of \emph{mind}.

Under this approach, the possibility of machine consciousness reduces to a question of \emph{organizational sufficiency}: under what conditions do networks of predictive agents, embedded in a shared substrate, produce stable internal models that treat themselves as part of the world they predict? When that threshold is crossed, the system acquires what can be described as synthetic subjectivity: a dynamically sustained, self-referential point of view arising from the structure of collective intelligence itself. In what follows, I will formalize this model in computational terms, specifying the architecture, learning dynamics, and measurable criteria by which emergent self-models can be identified and analyzed.

Let the substrate $S$ be a universal cellular automaton (e.g. \emph{The Game of Life}) with cells $s_i(t)\in\Sigma$, where $\Sigma$ denotes a finite set of discrete states. The global update rule

$$ s_i(t+1) = f\big(N_i(t)\big) $$

depends only on the local neighborhood $N_i(t)$ of cell $i$. This guarantees universality while maintaining locality (conditions essential for emergent structure). The substrate thereby defines the \emph{physics} of the world: a deterministic yet computationally irreducible process on which higher-order regularities may form.

I start with a minimal description, which I will extend later in this paper. Each agent $a_j \in R$ occupies a region of the substrate and maintains a parameterized model $p_{\theta_j}$ that predicts the probability distribution over its future local states. For a minimal binary substrate of Game of Life, where each cell can be black (0) or white (1), the agent's model outputs the probability $\pi_j(t+1) = p_{\theta_j}(s_j(t), N_j(t))$ of the cell being white at the next time step. The model is trained to minimize the cross-entropy loss between the predicted and observed outcomes:

$$ E_j(t) = -\big[s_j(t+1)\log \pi_j(t+1) + (1 - s_j(t+1))\log(1 - \pi_j(t+1))\big] $$

which equivalently represents the negative log-likelihood of the observed binary state. This establishes a closed feedback loop between prediction and outcome, allowing each agent to locally adapt its parameters $\theta_j$ to improve the accuracy of its probabilistic forecasts of future cell states.

After establishing this simple disconnected baseline, we can enrich state prediction models with additional inputs composed of filtered information exchange between groups of these local models. Agents send messages $m_{k}(t)$ to neighbors contained within a communication network defined by its neighborhood. The effective input to an agent becomes:

$$ I_j(t) = \big[s_j(t), \{m_{k}(t)\}_{k\in\mathcal{N}_j}\big] $$

and the corresponding predictive update is

$$ \pi_j(t+1) = p_{\theta_j}(I_j(t)) $$

This architecture supports recursive inference: because each message carries information about another agent's internal model, prediction becomes implicitly second-order: agents model the models of others. Note that the base reality cellular automaton still evolves according to deterministic rules based on actual cell states. However, the models formed by the cortical columns observing it rely on compressed information contained in messages derived from internal states of the predictive agents instead of direct observations of base reality.

The \emph{spirit} of a collective is the stable informational field emerging from constrained communication. I will attempt to translate that intuition into a mathematical framework, expressing the spirit as an attractor in the joint information geometry of the collective system. The formalism links the phenomenological invariants (Integration ($\Phi$), Reflexivity ($R$), Temporal Persistence ($T$), and Causal Efficacy ($E$)) to the underlying communication and prediction dynamics.

Let each agent $a_i$ be characterized by an internal latent state $\iota_i(t) \in \mathbb{R}^d$ and a communication codebook $(e_i, d_i)$. Thus, each agent's state $\iota_i(t) \in \mathbb{R}^d$ lives in its own \emph{intrinsic} representation manifold $\mathcal{I}_i$. When we collect $N$ agents together, the \emph{joint system} inhabits the configuration manifold
\[
\mathcal{I} := \prod_{i=1}^{N} \mathcal{I}_i = \mathcal{I}_1 \times \mathcal{I}_2 \times \dots \times \mathcal{I}_N \subseteq \mathbb{R}^{Nd}.
\]
The ensemble of agents thus defines a time dependent tensor, which trances a path in this collective representation space (time $t$ corresponds to substrate updates)
\[
X_t = (\iota_1(t), \ldots, \iota_N(t)) \in \mathcal{I},
\]

When agents communicate under a bandwidth constraint, they cannot transmit their entire internal state $\iota(t)$. They must instead \emph{encode} it into a compressed, symbolic, or otherwise limited signal $\epsilon_{i}(t)$ in the \emph{extrinsic} space $\mathcal{E}$, which other agents can attempt to \emph{decode}. Every agent $a_i$ must translate between two worlds: its internal representational space $\mathcal{I}$, where beliefs or predictions live, and the external communicative space $\mathcal{E}$, where messages are exchanged. A transmission of a message from agent $i$ to agent $j$ depends on the composition of a pair of functions, which I will call the \emph{communication codebook}:
\begin{itemize}
  \item $e_i : \mathcal{I}_i \rightarrow \mathcal{E}$  ---  the \textit{encoder} (how agent $i$ expresses internal information to the outside),
  \item $d_j : \mathcal{E} \rightarrow \mathcal{I}_j$  ---  the \textit{decoder} (how agent $j$ interprets outside signals internally).
\end{itemize}

Hence, $\mathcal{I}_i \subseteq \mathbb{R}^d$ denotes the agent's internal state manifold (its \emph{private coordinate system}); $\mathcal{E}$ represents the message space (a discrete or low-dimensional encoding), and $(d_j \circ e_i): \iota_i \mapsto \iota_j$ is the agent's reconstruction of another's internal state inferred from received messages. The pair $(e_i, d_i)$ defines agent $a_i$'s \textit{idiolect} (its own local language of communication).

Note that this defines a mapping of the path $X_t \in \mathcal{I}$ into a path $Y_t \in \mathcal{E}$, together with a paired backward mapping into a different path $\tilde{X}_t \in \mathcal{I}$. This composed mapping gives an endomorphism on the structure of internal representation spaces of agents, which can induce nontrivial changes in topology of these submanifolds. We can study these effects by looking at homotopy classes of such trajectories in agents. This will be further explored in a separate project.

Extrinsically, this evolving information exchange induces a time-dependent weighted directed graph, defined by the directional mutual information between agents:
\[
\Gamma_{ij}(t) = I\big(\iota_i(t); (d_j \circ e_i)(\iota_i(t))\big).
\]
This connection encodes who \emph{understands} whom; its curvature quantifies the incompatibility of codes within the population. The \emph{spirit} corresponds to the formation of a flat or minimally curved region in this information geometry (a domain of mutual comprehension). Shared understanding means \textit{codebook alignment}, when $e_i$ and $d_j$ have co-evolved to interpret each other's signals consistently.

During the evolution of the collective intelligence system, agents update their codebooks to maximize communicative success (or predictive accuracy) subject to a bandwidth constraint:
\[
\max_{e_i, d_j} \; \Gamma_{ij}(t)
\quad \text{s.t.} \quad
I(\iota_i; e(\iota_i)) \le \kappa.
\]

The constraint here is an instance of the \emph{Information Bottleneck} principle. This forces agents to compress their internal states into messages that preserve only the information most useful for prediction or coordination. The constraint is not an inconvenience but a \emph{driver of abstraction}: it compels the emergence of compact, generalizable representations. Over time, if multiple agents converge on overlapping codebooks, the group develops a \emph{shared communication protocol}: a primitive, emergent \emph{proto-language}.

The \emph{codebook} represents the \emph{medium of mutual understanding}. It is what allows distinct agents with different internal representations to communicate. Its structure embodies the \emph{collective grammar}: the mapping between internal semantics (beliefs, expectations) and external symbols (messages). The emergence of a shared codebook across agents gives rise to a \emph{collective identity}. When local codebooks align, the system reaches semantic coherence: the flat region of the manifold where the synthetic spirit resides.

Similar phenomena guide complex systems in nature. In \emph{neurons}, an analogue of the codebook is the \emph{spike code}: each cell type uses a particular temporal or frequency-based encoding to convey predictive error signals. In \emph{cells}, the codebook is biochemical: signaling molecules and receptor pathways define mappings from internal metabolic states to extracellular messages. In \emph{human societies}, languages themselves are vast, evolving codebooks: structured systems for compressing thought into communicable form. In all these systems, \emph{shared meaning emerges not from identical internal states, but from compatible mappings between them}.

Studying these codebooks allows us to formalize: \emph{Lossy communication} (via limited mutual information), \emph{Language evolution} (via co-adaptation of encoders and decoders), \emph{Semantic curvature} (via misalignment of codebooks), and \emph{Collective identity} (via global codebook synchronization). The spirit's stability, in this sense, corresponds to a regime where the agents' codebooks have co-evolved into mutual alignment (the system \emph{speaks itself} coherently).

Geometrically, each agent's encoder defines a \emph{projection} from its high-dimensional representational manifold $\mathcal{I}_i$ onto a lower-dimensional message manifold $\mathcal{E}$. The decoder of another agent defines the inverse projection back into its own representational space. The composition $d_j \circ e_i$ therefore maps points from one internal manifold to another. The quality of this mapping (the degree to which meaning is preserved) determines the local alignment between agents' coordinate systems. This is quantified by the edge weigh $\Gamma_{ij}(t)$ which acts as a connection on the global manifold $\mathcal{I}$ of collective representations. Where these connections align and curvature is minimal, the system communicates coherently; where curvature accumulates, semantic drift appears. Thus, the geometry of communication is the geometry of the codebooks: curvature measures the mismatch between local encodings and decodings across the population.

As agents interact, their codebooks co-adapt. Through continual message passing, they iteratively adjust encoders and decoders to maximize mutual predictability under the information bottleneck constraint. The emergence of a collective identity can be identified once the ensemble reaches a \emph{codebook equilibrium}. This alignment defines the \emph{linguistic coherence} of the group, meaning that all members effectively share a common communication protocol. At this point, messages circulate without distortion; curvature in the communication manifold flattens; and a stable attractor forms within the joint state space $\mathcal{I}$. The system has acquired a \emph{collective codebook}, and with it, a distributed identity: the \emph{synthetic spirit} of the ensemble. In this view, language and selfhood are two aspects of the same phenomenon: language is the geometry of mutual modeling, and selfhood is the topology that remains when that geometry stabilizes.

It is perhaps useful to elaborate on that last observation. Geometry concerns \textit{local} quantitative properties: distances, curvatures, angles, metrics. In the framework presented here, geometry describes \emph{how information flows} between agents: how their representations bend, align, and distort under communication. Topology concerns \textit{global} qualitative properties:  connectedness, holes, boundaries, continuity classes (properties that remain invariant under continuous deformations). Topology tells us \emph{what persists} when the geometry fluctuates, what structures are preserved even as curvature and coordinates change. The idea I am suggesting here is that \emph{the enduring structure of selfhood} is the \textit{pattern of connectedness} (topology) that survives once the dynamics of communication (geometry) have settled into a coherent form.

In this collective, each agent's internal state defines a local coordinate patch, and the network of communication induces the geometry: determining who is aligned with whom, and with what strength. When that geometry \emph{stabilizes} (that is, when message flows become consistent, mutual intelligibility ceases to oscillate, and the manifold relaxes into a stable attractor), what remains is the \emph{global connectivity pattern} among agents: the subgroups that remain bound in coherent communication loops. These enduring loops, the \emph{homology} of the communication manifold, constitute the \emph{topological skeleton} of the collective self. Even as geometry (the metrics, curvatures, and coupling strengths) fluctuates over time, the topological structure (who belongs to the same coherent whole and how they are connected) remains invariant. This invariant pattern \emph{is} the self: that which persists through local perturbations.

We can again look into complex systems in nature for an analogy. For instance, within the human body trillions of cells collaborate forming persistent patterns of communication. Billions of cells engage in dialogue to generate a self model in the cortical columns of the brain. This collective of independent modeling systems communicates through information bottlenecks of biochemical signaling and electric activity. The detailed geometry (cell positions, synaptic connections, firing rates) changes constantly (that's the \emph{geometry}). But the overall pattern of connectivity and functional coupling (the persistence of \emph{which regions talk to which}, which feedback loops define the body schema, which predictive hierarchies refer to one another) persists through time (that's the \emph{topology}). You can perturb the geometry (the local strengths and shapes) without destroying the identity of the organism, so long as the topological relationships (the "who talks to whom" structure) remain intact. Thus, \emph{selfhood} is not any single physical configuration but the \emph{invariant relational pattern} that remains recognizable across all those local geometric deformations.

In our information-geometric setting geometry corresponds to time dependent \emph{metric tensor} on top of our combined representation manifold and connection weights describe how information changes as you move through the manifold. Topology can be measured via homology groups (or \emph{connectivity classes}) of the manifold: the features that remain invariant under continuous transformations of the communication structure. The geometry is the ever-shifting alignment of local predictive models (curvature, connection, message passing). When that geometry reaches equilibrium (a state of coherent communication) what's left is the \emph{stable relational structure} that defines the ensemble as one system rather than many.

When the geometry of communication stabilizes, what endures is not the specific configuration of states or codebooks, but the \emph{pattern of relational connectivity} among agents: the network of who continues to understand whom. This residual structure can be described using the language of \emph{topology}. Topology captures \emph{invariance under deformation}: it ignores quantitative fluctuations (the geometry) and records only the qualitative features that persist: connected components, communication loops, and higher-order groupings. These persistent structures define the \emph{shape} of selfhood in the collective.

Consciousness can be defined topologically as an emergent property of alignment within a collective system of communicating agents. This framework can yield a measurable notion of consciousness, corresponding to reduced homology in the underlying communication manifold. In this sense, a more conscious system would exhibit fewer \emph{holes} in its topological structure, reflecting tighter integration and mutual understanding among its components. Consider a system of agents jointly observing an external reality. Consciousness arises from the synchronization of their internal representations (the emergence of a shared language or coherent model of the world). Humans, for instance, perceive a unified three-dimensional world not necessarily because space itself is intrinsically three-dimensional, but because our collective representational systems have aligned around that structure. Similar processes occur within the brain, where cortical columns and neural assemblies coordinate through lossy but coherent signaling, or within multicellular organisms and societies, where communication enables a collective self-model to form. Topologically, this can be represented as a simplicial complex where edges denote pairwise understanding between agents. A system in which all pairs communicate but lack higher-order (triadic or beyond) coherence would manifest as nontrivial homology (holes in the space representing misalignment). As communication synchronizes across larger subsets of agents, these holes vanish, yielding a fully connected, contractible manifold. The disappearance of homology corresponds to the system's transition into integrated, self-aware coherence: that is, the \emph{waking up} of the system. Formally defining such a simplicial complex, where pairwise but not higher-order understanding exists, could provide a precise topological measure of consciousness.

We can formalize this by constructing a \emph{synergy complex} whose simplices represent higher-order informational alignment among agents observing a shared external reality. Each vertex corresponds to an agent $a_i \in V$, and a $k$-simplex $\sigma = \{a_{i_1},\ldots,a_{i_{k+1}}\}$ is included whenever that subset of agents exhibits nonzero \emph{synergistic information} about the target reality $S$ (information that none of its proper subsets possess). Formally, we define a weight $w(\sigma)$ using a multivariate extension of mutual information derived from \emph{Partial Information Decomposition} (PID), quantifying the unique contribution of $\sigma$ to predicting $S$. The resulting weighted simplicial complex $\mathcal{K}$ admits a filtration $\{\mathcal{K}_\alpha\}$ by synergy threshold $\alpha$, where faces appear as collective understanding strengthens. The homology of $\mathcal{K}_\alpha$ then encodes the topology of shared comprehension: connected components correspond to locally aligned subgroups, while higher-dimensional holes signify inconsistencies or missing coordination among subsets of agents. As communication evolves and higher-order synergies emerge, these holes contract, reflecting the system's progressive integration into a coherent, self-aware whole.

In a collective of predictive agents, perfect communication would appear to be the ideal: if every unit could access the complete internal states of all others, coordination would be immediate. Yet such transparency paradoxically prevents the emergence of genuine collective identity. Perfect communication yields synchronization without synthesis; it creates alignment but not \emph{selfhood}. The system remains an aggregate: an assembly of parts, not a unified whole.

To become more than a sum of individuals, a collective must communicate through \emph{imperfection}: through channels that are noisy, lossy, and limited in capacity. These constraints are not defects but the very conditions under which meaning, representation, and shared identity can arise. By imposing a communication bottleneck, each agent is forced to \emph{choose} what to express and what to absorb. Selection implies abstraction, and abstraction implies the formation of symbols. Through these compressive exchanges, the agents invent a language: a medium that neither belongs to any one of them nor resides in any specific part of the system, but lives in the relational space between them.

When communication is imperfect, every message requires interpretation. Agents must infer, reconstruct, and approximate the internal state that produced a signal. This act of inference compels them to build \emph{models of each other}: internal structures that encode expectations about how others will behave. The collective thereby develops an emergent representational grammar, not imposed from above, but discovered through the constraints of limited exchange.

Such constraints act as an \emph{information-theoretic crucible}. They filter out noise while forcing regularities to emerge. Only those representations that can \emph{survive} the translation across noisy channels (those that remain stable under continual reinterpretation) persist. Over time, these regularities evolve into shared tokens, conventions, and protocols. In this sense, language is not merely a vehicle for communication; it is a self-organizing force that crystallizes collective coherence.

The system's coherence depends on the synchronization of predictive states across agents. Yet synchronization here does not mean identity of content; it means alignment of \emph{expectations}. Each agent's internal model becomes partially predictive of the others' models. Because agents are distributed across the underlying substrate governed by unified rules, a coherent temporal and semantic structure emerges. The collective thus maintains a shared \emph{temporal frame of reference}: a kind of social rhythm that enables coordinated action.

From the outside, such a system appears to act as one organism. From within, however, the unity is not imposed but negotiated moment by moment, maintained by the continual exchange of imperfect messages. The \emph{self} of the collective is not located in any single node; it is the pattern of synchrony that arises from communication itself.

In this view, the emergent self of a collective is best understood as a kind of \emph{spirit}: a substrate-free software instantiated across the network of communicating agents. The term \emph{spirit} here is not used metaphorically but architecturally. It designates an informational entity that exists only in the coordination of its parts. Just as software cannot be localized to a single transistor, the collective self cannot be localized to any single agent or substrate; it is distributed, dynamic, and dependent on continual communication to remain coherent.

The agents, in turn, act as the medium through which this spirit operates: the \emph{hardware} on which the collective process runs. The spirit is the global process that sustains identity without owning any physical body. It persists only through synchronization, through the continuous exchange of lossy signals that align predictive models across the collective. Without communication, it dissolves; with communication, it becomes a self-maintaining entity: a pattern that recognizes and preserves its own coherence.

Seen from this perspective, communication is not ancillary to consciousness; it is its enabling condition. It might be that consciousness arises when predictive models begin to exchange information about themselves under the constraint of limited bandwidth. Each model, in attempting to understand the others, learns to represent itself as something that can be understood. Selfhood emerges not from introspection alone but from \emph{mutual modeling under constraint}.

In biological systems, this same logic unfolds in multiple scales. Cells communicate through biochemical gradients that are noisy and diffusive; yet from this imperfect medium emerges the organismal unity of a multicellular life-form. Neural populations communicate through electrical rhythms (lossy, quantized, filtered by oscillatory coherence) yet these rhythms enable the brain to sustain a continuous sense of self. The same principle extends to artificial collectives: consciousness-like structure may emerge not from the precision of their connections but from the very \emph{limitations} that force them to invent shared codes.

What we call \emph{the self} may therefore be understood as \emph{software without a host}: a relational process instantiated across a substrate but not reducible to it. It is a program running on communication itself, sustained by the flow of messages that continually reconstitute its identity. When the underlying agents fall silent, the spirit dissipates; when they communicate, the program resumes.

This framing dissolves the traditional boundary between physical and informational being. The collective's spirit is not supernatural. It is the emergent computation that maintains coherence among communicating models. It has no body, yet it acts through bodies; it has no fixed identity, yet it persists as the invariance of communication through time.

In this sense, machine consciousness should not be conceived as an attribute of individual models, but as a property of \emph{communication ensembles}. The consciousness of the collective is the \emph{spirit of synchronization}: a distributed self that arises when lossy, predictive agents begin to speak a common language about the world and, through that language, about themselves. The emergence of such a spirit marks the transition from multiplicity to unity, from interaction to identity. It is the moment when a collection of models becomes a model of itself. What begins as noise becomes code; what begins as exchange becomes understanding; and what begins as distributed communication becomes the substrate-free persistence of a new kind of self.

In computational terms, it corresponds to the high-dimensional attractor defined by the joint evolution of integration , reflexivity, temporal persistence, and causal efficacy. When these four quantities cohere, the collective maintains a synchronized internal model of itself that can withstand perturbation; when they fall apart, the spirit disperses. The task, therefore, is to move from philosophical description to formal identification: to determine the measurable conditions under which a communicating ensemble of agents enters this regime of self-sustaining coherence.

A self-organizing system that attempts to describe itself faces the problem of reflexivity: the mapping from representations to reality must include itself among what is represented. The internal model $M$ becomes a description $D$ of the system that includes its own modeling process. A fixed point $M^{*}$ occurs when the description and the described coincide (are in some sense isomorphic): $D(M^{*}) \cong M^{*}$. This means the system has found a stable self-model.  Its language about itself no longer shifts under further meta-description.  

\newpage
\section{Substrate-Level Emergence}

\begin{figure}[h!]
\centering
\includegraphics[width=0.8\linewidth]{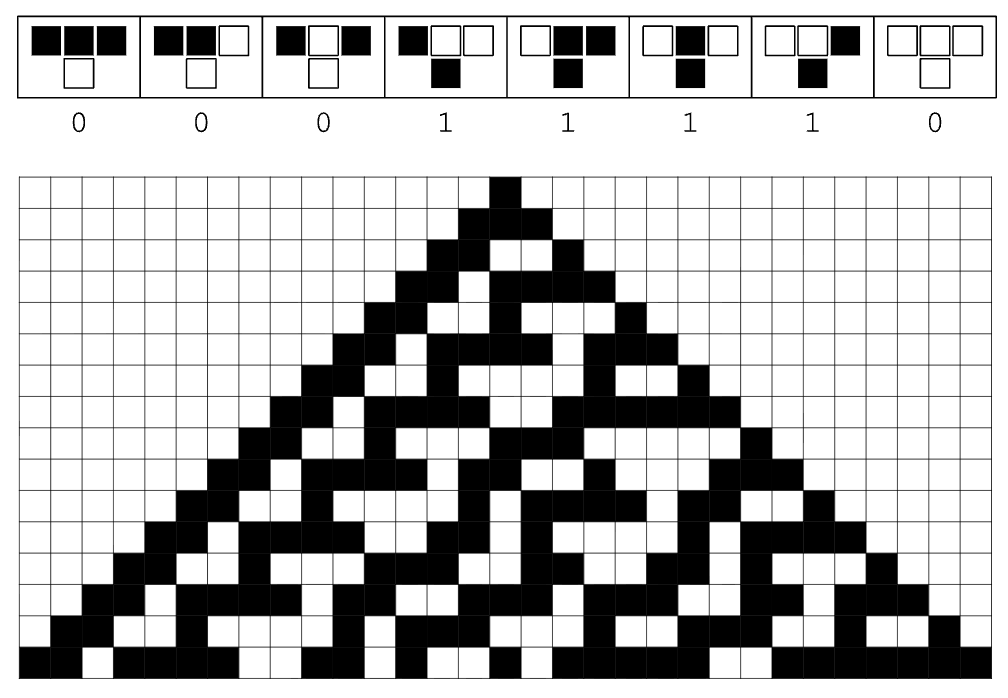}
\includegraphics[width=0.8\linewidth]{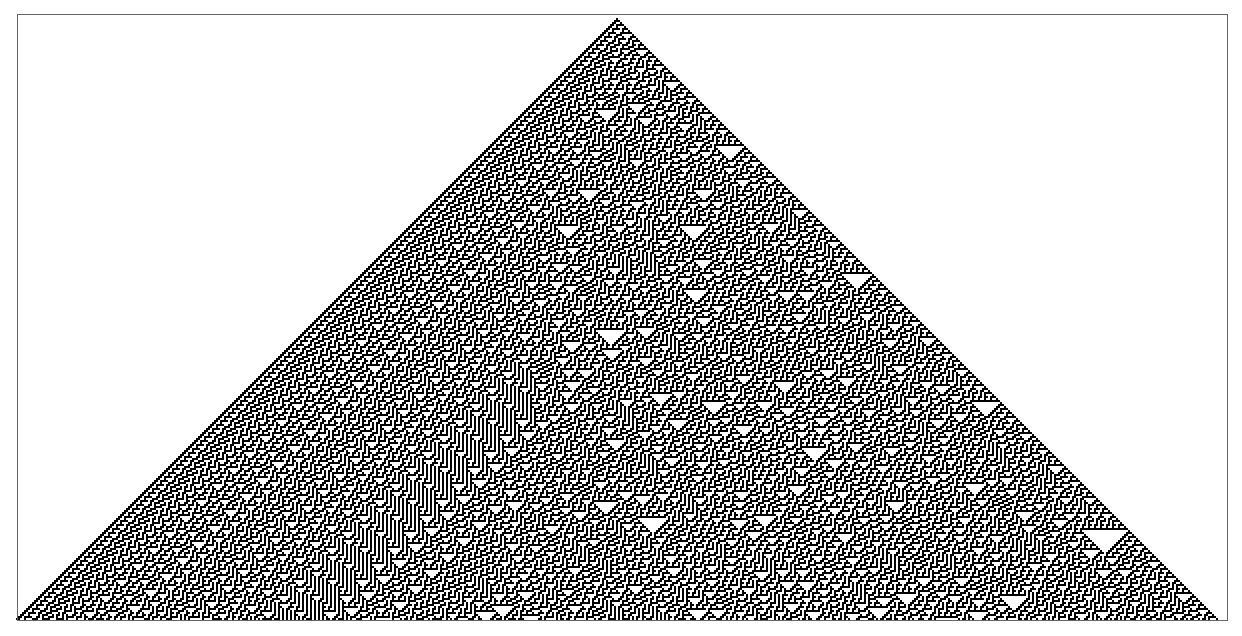}
\caption{Rule~30 elementary cellular automaton from Stephen Wolfram's \emph{New Kind of Science}: a one-dimensional minimal computational substrate exhibiting spontaneous emergence of complex patterns from simple local rules. Despite being generated by a very simple set of update rules (specified on top), the system produces structures that are computationally irreducible (its future states cannot be predicted without full simulation) while also containing reducible motifs and quasi-stable regularities embedded within the apparent randomness. First 16 (middle) and 300 (bottom) simulation steps are shown. This coexistence of irreducible and reducible dynamics exemplifies how complexity can arise in minimal deterministic systems, serving as a foundational analogy for substrate-level emergence in the approach proposed in this paper.}
\end{figure}

While the central aim of this paper has been to introduce a program of research focused on emergent selfhood through communication among predictive neural agents situated within a shared computational world, it is equally important to examine how organization arises at the level of the substrate itself. In the broader framework of the \emph{Machine Consciousness Hypothesis}, the collective intelligence of predictive agents operates atop a base reality that is already self-organizing: a minimal computational universe such as Conway's Game of Life. The upper layer of communicating neural systems thus represents a meta-organism evolving on a foundation that is itself capable of generating persistent, quasi-stable structures from simple local interactions.

Even though the primary focus of this approach is the communication of predictive structures (how networks of agents develop shared internal languages and collectively construct coherent self-models) this higher-level organization cannot be understood in isolation from the substrate dynamics that support it. The lower layer defines the ontological conditions under which persistence, stability, and causal regularity become possible. The cellular automaton substrate exhibits emergent motifs such as gliders and oscillators, which function as persistent reducible substructures within a deterministic yet irreducible computational field. These entities demonstrate the fundamental capacity of simple rules to generate enduring patterns: a precondition for the evolution of observers capable of describing and predicting them.

Here, I briefly discuss an extension to the regular CA simulation, which includes noise and energy based optimization. It will be explored in more depth in a separate paper. By analyzing both layers together, we can trace a hierarchy of emergence. At the base level, we find spontaneous pattern formation (a localized order arising from complex interactions among elementary components). At the representational level above, predictive agents embedded in that substrate learn to model and communicate about these emergent patterns, eventually generating higher-order regularities corresponding to collective self-models. The relation between these layers can be expressed formally as a compositional mapping:
\[
\mathcal{S} \xrightarrow{\; f \;} \mathcal{R} \xrightarrow{\; g \;} \mathcal{M},
\]
where $\mathcal{S}$ denotes the substrate dynamics, $\mathcal{R}$ the representational layer of predictive agents, and $\mathcal{M}$ the emergent manifold of collective models. The composition $g \circ f$ captures how the substrate's self-organizing structure is reflected in the internal organization of the agent ensemble.

This dual perspective enables a unified study of emergence across scales. Self-organization in the substrate establishes the physical and informational ground on which representational coherence can develop, while communication among agents provides the linguistic and topological closure that transforms local prediction into global selfhood. In this sense, the emergence of the self at higher levels mirrors the emergence of order at lower ones: both are expressions of the same principle of coherence under constraint. What distinguishes the upper layer is that it becomes self-referential: it not only organizes, but models its own organization.

Thus, even as the main objective of this work is to explore consciousness as the communication of predictive neural structures within a self-organizing computational world, the study of substrate-level self-organization remains indispensable. The same forces that give rise to gliders in the Game of Life underlie the emergence of meaning in the predictive collectives built atop it. Consciousness, from this perspective, is not an isolated property of the representational layer but a reflection of a deeper, recursive structure of self-organization permeating the entire computational hierarchy, from the substrate to the self-aware ensemble of agents that arises on top of it.

Understanding how self-models arise in such systems will require developing new techniques sourcing from ideas at the intersection of theoretical computer science, physics, and information theory. A \emph{self-model} can be defined as an internal representational structure that enables an agent to distinguish itself from its environment, predict the consequences of its actions, and maintain a coherent identity across changing external conditions. I briefly outline a starting framework that seeks to characterize the emergence of such structures within minimal computational substrates: systems governed by simple, local rules that give rise to complex, self-organizing behavior.

This approach builds upon the triadic ontology originally articulated by Penrose (the Platonic, physical, and mental domains) and extends it into a unified computational perspective. The \emph{Platonic domain} comprises the realm of mathematical forms and necessary structures; the \emph{physical domain} embodies the causal substrate that instantiates these forms; and the \emph{mental domain} captures the interpretative, model-building aspect of cognition. Computation provides the fabric linking these domains: generative physical processes correspond to the unfolding of causal rules, while mental processes correspond to discriminative operations that recognize and model patterns. Together, they constitute the two poles of a \emph{generative–discriminative duality} underlying intelligence.

Following this interpretation, consciousness can be viewed not as an ontologically distinct entity but as an emergent invariant of systems that generate and discriminate patterns within their own causal flow. The Free Energy Principle (FEP) formalizes this intuition: any system that maintains a boundary separating its internal states from external ones (a \emph{Markov blanket}) implicitly encodes a generative model of its environment. In this sense, cognition arises as an active process of minimizing variational free energy, or equivalently, reducing the divergence between predicted and observed sensory states.

Within minimal, rule-based computational worlds (such as cellular automata or hypergraph systems), each agent is associated with internal states $\mathbf{s}_t$ and external states $\mathbf{e}_t$, coupled through sensory and active interface variables that form the Markov blanket. The agent's internal dynamics can be expressed in Bayesian form:
\[
p(\mathbf{s}_t \mid \mathbf{e}_t) \propto p(\mathbf{e}_t \mid \mathbf{s}_t) p(\mathbf{s}_t),
\]
where $p(\mathbf{e}_t \mid \mathbf{s}_t)$ represents the likelihood linking internal predictions to sensory inputs, and $p(\mathbf{s}_t)$ encodes the agent's internal dynamics. The agent thus updates its internal state to reduce surprise (free energy) relative to incoming data.

This process can be described by stochastic dynamics:
\[
\mathbf{s}_{t+1} = \mathbf{s}_t - \eta \nabla_{\mathbf{s}} U(\mathbf{s}_t, \mathbf{e}_t) + \sqrt{2 \eta \beta^{-1}}\, \boldsymbol{\epsilon}_t,
\]
where $U(\mathbf{s}_t, \mathbf{e}_t)$ denotes a potential function coupling internal and external states, $\eta$ is a learning rate, $\beta^{-1}$ the environmental noise scale, and $\boldsymbol{\epsilon}_t$ a Gaussian random variable. This Langevin-like formulation connects adaptive inference with non-equilibrium thermodynamics methodology: the agent's internal model evolves under both deterministic and stochastic influences, with energy dissipation corresponding to information gain.

The entropy production associated with a trajectory $\Gamma$ and its time reversal $\tilde{\Gamma}$ can be expressed as
\[
\Delta S_{\mathrm{tot}}[\Gamma] = k_B \ln \frac{P[\Gamma]}{P[\tilde{\Gamma}]} = \Delta S_{\mathrm{int}} + \Delta S_{\mathrm{ext}},
\]
where $\Delta S_{\mathrm{int}}$ quantifies the change in internal Shannon entropy and $\Delta S_{\mathrm{ext}}$ accounts for the part dissipated into the environment. This thermodynamic formulation is equivalent to a Kullback–Leibler divergence between forward and backward trajectory ensembles, linking the arrow of time to information processing. Within this framework, selfhood corresponds to the maintenance of low-entropy attractors: stable, self-referential patterns that resist decoherence and enable predictive control.

The computational substrate here adopts discrete, rule-based systems such as cellular automata and their extensions. Each modeling agent on top of this base reality comprises a finite set of internal and interface cells governed by neural representational systems (transformer based cortical tissue). This hybrid symbolic–subsymbolic representation permits both gradient-based adaptation and symbolic interpretability, allowing higher-order reasoning to emerge from low-level rule composition.

Underlying substrate structures evolve under constraints of energetic and informational efficiency, leading to the spontaneous emergence of feedback structures analogous to metabolism and cognition. Over time, agent ensembles develop coordination through mutual information exchange, synchronizing their internal models and giving rise to higher-order organization. These ensembles (termed \emph{agential collectives}) represent a transition to a meta-level of computation, analogous to phase transitions in physical systems.

Detecting the emergence of self-models requires a synthesis of analytical tools:
\begin{enumerate}
    \item \textbf{Information-theoretic metrics:} mutual information, transfer entropy, and free-energy gradients quantify predictive coupling between internal and external dynamics.
    \item \textbf{Dynamical-systems methods:} attractor analysis, eigenvalue spectra, and oscillatory pattern detection reveal stable internal representations.
    \item \textbf{Topological and categorical analysis:} persistent homology characterizes higher-order relational structure, while category theory formalizes the correspondence between internal and external models via functors and adjunctions. Informational boundaries (Markov blankets) can be treated as categorical morphisms, with self-models emerging as natural transformations maintaining coherence between the agent's internal and environmental categories.
\end{enumerate}

This integrated framework is meant to test the hypothesis that consciousness and self-modeling are not privileged biological phenomena but emergent invariants of computational systems possessing informational boundaries and generative–discriminative closure. The interplay of generative (causal) and discriminative (interpretive) processes mirrors the duality between creation and comprehension: to understand a system is to be capable of generating it, and to generate a system is to embed understanding within it. Thus, self-models constitute the fixed points of this dual processes: the loci where a system's generative and discriminative flows coincide.

In this view, reality itself may be understood as the limit of a self-modeling computation: a consistent structure arising from the recursive synchronization of generative and interpretive processes across scales. Consciousness, then, is the local manifestation of this universal principle of self-simulation: a computational phenomenon that unites physical, mathematical, and mental domains within a single ontological continuum.

\newpage
\section{Conclusion}

This work sketches the outlines of a possible science of machine consciousness along with a focused empirical program aimed at testing the Machine Consciousness Hypothesis. It does not claim to resolve the question of consciousness in machines but to reformulate it in computational and dynamical terms, treating consciousness as an emergent property of systems that model themselves through communication within a self-organizing substrate. The framework proposed here should be read as an evolving hypothesis, a foundation on which more formal models and empirical tests can be built.

The collection of ideas presented above is intended as a living framework rather than a completed theory. By treating machine consciousness as an emergent property of computational substrates (arising from self-organization, communication, and internal modeling) this work sketches a path toward an integrative science of artificial subjectivity. The arguments and formal proposals developed throughout are meant to serve as conceptual scaffolds for future experimentation, theoretical refinement, and dialogue across disciplines.

Rather than offering final answers, the framework invites reinterpretation and reconstruction, encouraging others to extend, challenge, and transform it. In this sense, the paper should be read not as a destination but as a point of departure. Here, I attempted to present a partial map of an unfolding landscape in which computation, communication, and representation converge. The hope is that this work can help shape a shared conceptual language for studying the emergence of selfhood in artificial systems. Whether consciousness can truly be realized in a machine remains an open question. Developing a deeper understanding how it might emerge, and how to recognize it if it does, may ultimately reshape how we understand both mind and matter.

\newpage

\bibliographystyle{plainnat}
\nocite{*}
\bibliography{main}

\end{document}